# Bayesian exponential family projections for coupled data sources

**Arto Klami, Seppo Virtanen, Samuel Kaski**
Aalto University School of Science and Engineering
Department of Information and Computer Science
Helsinki Institute for Information Technology HIIT
P.O.Box 15400, FI-00076 Aalto, Finland

## Abstract

Exponential family extensions of principal component analysis (EPCA) have received a considerable amount of attention in recent years, demonstrating the growing need for basic modeling tools that do not assume the squared loss or Gaussian distribution. We extend the EPCA model toolbox by presenting the first exponential family multi-view learning methods of the partial least squares and canonical correlation analysis, based on a unified representation of EPCA as matrix factorization of the natural parameters of exponential family. The models are based on a new family of priors that are generally usable for all such factorizations. We also introduce new inference strategies, and demonstrate how the methods outperform earlier ones when the Gaussianity assumption does not hold.

## 1 INTRODUCTION

Principal component analysis (PCA) is arguably the most popular machine learning algorithm, used for example as the standard choice for dimensionality reduction. When PCA is interpreted as a probabilistic latent-variable model (Tipping and Bishop, 1999), it becomes obvious that it assumes both the latent variables and the observations to be Gaussian. This assumption is ultimately rather restrictive, which explains the recent renewed interest in PCA.

One of the main research directions has been to relax the Gaussianity assumption to better fit domains where data is not continuous-valued. The exponential family variant of PCA (EPCA; Collins et al., 2002) introduced a way of taking the data distribution into account. It was followed by a semi-parametric formulation applicable to even more flexible distributions (Sajama and Orlitsky, 2004), by more efficient algorithms converging into the global optimum (Guo and Schuurmans, 2008), and supervised PCA (both exponential family and standard; Yu et al. 2006; Guo 2009). Bayesian exponential family PCA takes the approach to the next level, by including a fully probabilistic model that needs not assume deterministic latent variables. Mohamed et al. (2009) made a straightforward assumption of Gaussian priors which is suboptimal but hard to remove in practice since the latent variables do not follow any standard distribution for general exponential families. We introduce a novel regularizing prior for EPCA models that removes some of the problems of the Gaussianity assumption, and present inference solutions compatible with the prior.

In an abstract and compact form, the PCA problem is simply a matrix decomposition. The $N \times D$ data matrix $\mathbf{X}$ is decomposed as $\mathbf{X} = \mathbf{UV}$, where $\mathbf{U}$ and $\mathbf{V}$ are of low rank and the scales have been incorporated into either matrix. EPCA makes this decomposition in the space of the natural parameters of element-wise exponential family distributions. That is, we assume that each element of $\mathbf{X}$ has been generated independently from an exponential family distribution with parameters collected into $\mathbf{\Theta}$, while $\mathbf{\Theta}$ itself is factorized as $\mathbf{\Theta} = \mathbf{UV}$.

The resulting model family contains as special cases extensions of several useful projection methods, in addition to standard PCA. We consider joint analysis of two (or in general more) data sources, demonstrating how Bayesian exponential family variants of supervised EPCA (Guo, 2009), partial least squares, and canonical correlation analysis can be obtained using the same basic formulation. The proposed methods extend naturally the recent literature on probabilistic variants of these methods (PLS: Gustafsson, 2001; Nounou et al., 2002; CCA: Bach and Jordan, 2005; Klami and Kaski, 2007) in the same way as the EPCA approaches extend probabilistic PCA. Moving from EPCA to models of coupled sources poses some techni-cal challenges for the inference. We introduce a practical approximate Bayesian inference technique that effectively solves many of those for our proposed prior.

We will first recap the general form of exponential family projection models, and proceed to introduce in detail the assumptions that result in the multi-view models of coupled data sources. Then we introduce ways of defining priors for the models, and present inference algorithms. Finally, the models are demonstrated to outperform their rivals in a number of experiments using both artificial and real data.

## 2 EXPONENTIAL FAMILY PCA

A vectorial random variable $\mathbf{x} \in \mathbb{K}^D$ (where $\mathbb{K}$ is a suitable subset of the real-space, such as $\mathbb{Z}$ or $\mathbb{R}_+$) in the exponential family follows the distribution

$$p(\mathbf{x}|\boldsymbol{\theta}) = \exp(s(\mathbf{x})^T\boldsymbol{\theta} + h(\mathbf{x}) - g(\boldsymbol{\theta})), \quad (1)$$

where the elements of $\mathbf{x}$ are assumed independent of each other. Here $\boldsymbol{\theta} \in \mathbb{K}^D$ represents the natural parameters of the distribution, $g(\cdot)$ is the log cumulant function that normalizes $p(\mathbf{x}|\boldsymbol{\theta})$ to be a valid distribution, $s(\cdot)$ are the sufficient statistics, and $h(\cdot)$ is a function of the data alone. We choose the natural exponential family by assuming $s(\mathbf{x}) = \mathbf{x}$. The function $g(\cdot)$ then defines the distribution: different choices lead to different exponential family distributions including Gaussian with unit variance, Bernoulli, Poisson, and exponential to name a few. For every member of the exponential family there exists a conjugate prior distribution for $\boldsymbol{\theta}$:

$$p(\boldsymbol{\theta}) \propto \exp(\boldsymbol{\lambda}^T\boldsymbol{\theta} - \nu g(\boldsymbol{\theta})).$$

Exponential family PCA is computed in the natural parameter space of the element-wise exponential family. The element $\mathbf{X}_{nd}$ is assumed to come from $p(\mathbf{X}_{nd}|\boldsymbol{\Theta}_{nd})$, independently for all elements. Thus the matrix $\mathbf{X}$ is assumed to come from distribution $p(\mathbf{X}|\boldsymbol{\Theta})$, where we further assume $\boldsymbol{\Theta} = \mathbf{UV}$ is of low rank; $\mathbf{U}$ and $\mathbf{V}$ have a low pre-specified number $K$ of columns and rows, respectively[1]. Essentially, we make the conditional independence assumption between elements of $\mathbf{X}$ given $\boldsymbol{\Theta}$, assuming that $\boldsymbol{\Theta}$ is flexible enough to capture the relevant structure in data. The rows of $\mathbf{U}$ can be interpreted as latent variables generating the data, and $\mathbf{V}$ as projections transforming the latent variables. With Gaussian distribution and Gaussian latent variables this would result in probabilistic PCA.

[1] For non-centered data one can include a separate rank-one mean parameter in the factorization, controlling the mean of each feature. We leave the mean parameter out to simplify the formulas.

Bayesian treatment of EPCA (Mohamed et al., 2009) requires prior distributions for the model parameters $\mathbf{U}$ and $\mathbf{V}$. We will discuss the priors in more detail in Section 4 after introducing a set of coupled data analysis models obtained as special cases of EPCA.

## 3 MODELS FOR COUPLED DATA

### 3.1 GENERAL FORM

Two data sets, $\mathbf{Y}_1 \in \mathbb{K}^{N \times D_1}$ and $\mathbf{Y}_2 \in \mathbb{K}^{N \times D_2}$, are coupled if the samples co-occur; each row of $\mathbf{Y}_1$ is paired with the corresponding row in $\mathbf{Y}_2$. By concatenating the two sources as $\mathbf{X} = \begin{pmatrix} \mathbf{Y}_1 & \mathbf{Y}_2 \end{pmatrix}$ we can write several projection methods for coupled data sources as EPCA of $\mathbf{X}$, that is, as factorizations of the form $\boldsymbol{\Theta} = \mathbf{UV}$, where certain elements of $\mathbf{V}$ are restricted to be zero. Many of the decisions in practical modeling, such as the choice of prior distributions and inference algorithm, are independent of such restrictions imposed on $\mathbf{V}$, and hence the unified framework helps in developing practical algorithms for various coupled data analysis tools.

### 3.2 SUPERVISED EPCA

Supervised PCA is the simplest model for coupled data. One of the sources, say $\mathbf{Y}_1$, is treated as the target variable, and the task is to find a low-dimensional representation of $\mathbf{Y}_2$ that helps in predicting the target. In the simplest case one obtains the solution directly as the EPCA of $\mathbf{X}$. The original SPCA formulation (Yu et al., 2006) as well as the supervised EPCA (Guo, 2009) follow this idea, the crucial difference being that the latter makes a suitable distributional assumption for discrete target variables. Due to the basic assumption of independence over the features, the model is written as

$$p(\mathbf{Y}_1, \mathbf{Y}_2, \mathbf{U}, \mathbf{V}) = p(\mathbf{Y}_1|\mathbf{U}, \mathbf{V}_1)p(\mathbf{Y}_2|\mathbf{U}, \mathbf{V}_2) \\ p(\mathbf{U})p(\mathbf{V}_1)p(\mathbf{V}_2),$$

where $\mathbf{V} = \begin{pmatrix} \mathbf{V}_1 & \mathbf{V}_2 \end{pmatrix}$ so that the columns are split according to the features in $\mathbf{X}$. No distinction is made between the features in $\mathbf{Y}_1$ and $\mathbf{Y}_2$, and hence the supervision is weak.

The predictive performance improves if one does not attempt to model $\mathbf{Y}_2$ perfectly; the ultimate task is to predict $\mathbf{Y}_1$ and the covariates $\mathbf{Y}_2$ should be modeled only to the degree they help in that. Rish et al. (2008) solved this by introducing a weighting for the generative parts,

$$p(\mathbf{Y}_1, \mathbf{Y}_2, \mathbf{U}, \mathbf{V}) = p(\mathbf{Y}_1|\mathbf{U}, \mathbf{V}_1)p(\mathbf{Y}_2|\mathbf{U}, \mathbf{V}_2)^\alpha \\ p(\mathbf{U})p(\mathbf{V}_1)p(\mathbf{V}_2),$$

where $\alpha$ controls the relative importance of modeling the two sources. A small value for $\alpha$ equals spending less modeling power on the covariates, resulting in increased predictive performance. The value is chosen for instance by cross-validation.

## 3.3 EXPONENTIAL FAMILY PLS

An alternative way of improving the predictive performance in supervised learning tasks is to allow the covariates to have structured noise that is independent of the target variable. This leads naturally to a classical linear supervised dimensionality reduction method of partial least squares (PLS) and its probabilistic variants (Gustafsson, 2001; Nounou et al., 2002). We extend the idea to the exponential family distributions.

The key idea in EPLS is that not all variation in $\mathbf{Y}_2$ is relevant for predicting $\mathbf{Y}_1$. As a generative model we restrict some of the components to only model $\mathbf{Y}_2$. By factoring $\mathbf{U} = \begin{pmatrix} \mathbf{U}_S & \mathbf{U}_2 \end{pmatrix}$ and $\mathbf{V}$ as

$$\mathbf{V} = \begin{pmatrix} \mathbf{V}_{S1} & \mathbf{V}_{S2} \\ \mathbf{0} & \mathbf{V}_2 \end{pmatrix},$$

where $S$ indicates shared variables, we can still write the model as $\mathbf{\Theta} = \mathbf{UV}$. The model complexity is governed by setting the ranks of the various parts. Denoting the rank of $\mathbf{U}_S$ by $K_S$ and the rank of $\mathbf{U}_2$ by $K_2$, the zeroes in $\mathbf{V}$ the make sure the last $K_2$ columns of $\mathbf{U}$ will have no effect on $\mathbf{Y}_1$. In more intuitive terms we have $\mathbf{\Theta} = \begin{pmatrix} \mathbf{\Theta}_1 & \mathbf{\Theta}_2 \end{pmatrix}$, where the parameters can equivalently be written as

$$\begin{aligned} \mathbf{\Theta}_1 &= \mathbf{U}_S \mathbf{V}_{S1} \\ \mathbf{\Theta}_2 &= \mathbf{U}_S \mathbf{V}_{S2} + \mathbf{U}_2 \mathbf{V}_2. \end{aligned} \quad (2)$$

This makes explicit the assumption that all variation in the target variable must come from the shared latent sources, while the covariates are created as an additive sum of the shared and source-specific variation.

We will later show in the experiments how EPLS requires less shared components for predicting $\mathbf{Y}_1$ than supervised PCA, which makes the model easier to interpret.

## 3.4 EXPONENTIAL FAMILY CCA

Going beyond mere prediction problems, a common task in analysis of coupled data is finding what is shared between the two data sources. This is a kind of data fusion task; the goal is to compress two data sources into a representation that captures the commonality between the two. The problem has traditionally been solved by canonical correlation analysis, or its kernelized variant, that have been applied to a range of practical problems such as extracting shared semantics of document translations (Vinokourov et al., 2003) and discovering dependencies between images and associated text to be used as preprocessing for classification (Farquhar et al., 2006). The probabilistic interpretation of CCA (Bach and Jordan, 2005) shows that the classical CCA implicitly assumes normal distribution. We remove this assumption and present a novel generalization of CCA to the exponential family that is useful, e.g., for analysis of text documents described as binary word occurrences or counts.

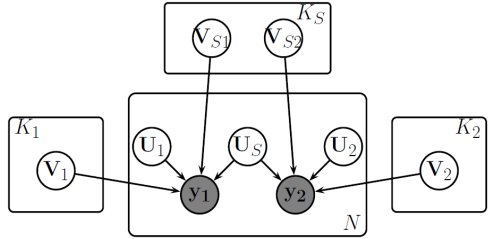

Figure 1: Graphical model for Bayesian exponential family CCA.

In the exponential family projection framework, ECCA is obtained by factoring the parameters as $\mathbf{U} = \begin{pmatrix} \mathbf{U}_S & \mathbf{U}_1 & \mathbf{U}_2 \end{pmatrix}$ and

$$\mathbf{V} = \begin{pmatrix} \mathbf{V}_{S1} & \mathbf{V}_{S2} \\ \mathbf{V}_1 & \mathbf{0} \\ \mathbf{0} & \mathbf{V}_2 \end{pmatrix},$$

following the presentation of Archambeau and Bach (2009). The notation is equivalent to (Klami and Kaski, 2008)

$$\begin{aligned} \mathbf{\Theta}_1 &= \mathbf{U}_S \mathbf{V}_{S1} + \mathbf{U}_1 \mathbf{V}_1 \\ \mathbf{\Theta}_2 &= \mathbf{U}_S \mathbf{V}_{S2} + \mathbf{U}_2 \mathbf{V}_2. \end{aligned}$$

Depending on the task we then analyze either the shared variables $\mathbf{U}_S$, implicitly marginalizing out the other parts, or either one of the source-specific ones. The full model is illustrated in Figure 1, to clarify the role of the various parts of $\mathbf{V}$.

## 4 PRIORS FOR EXPONENTIAL FAMILY PROJECTIONS

We will now turn our focus to the prior distributions given for the model parameters, which is an open problem for exponential family models in general and coupled-data models in particular. We propose a family of prior distributions that incorporates certain common choices as special cases, while being an efficient way of altering a compromise between conjugacy and flexibility in practical models.

Mohamed et al. (2009) extended the EPCA to a full Bayesian model, specifying priors directly for $\mathbf{U}$ and

**V**. This approach is conceptually simple and straightforward, but it is hard to determine which distributions to use. Mohamed et al. (2009) borrowed the assumption of normally distributed latent variables **U** from the Gaussian case, while taking **V** conjugate to the specific exponential family. Unfortunately that choice is poor for some exponential family distributions. For example, for the exponential distribution the domain of $\boldsymbol{\Theta}$ is the set of positive real numbers, which does not comply with the normal distribution.

Another intuitive alternative is to impose the prior on the product **UV**, instead of formulating separate priors for each variable. For **UV** we can easily choose a prior conjugate to the specific distribution of **X**, which makes the estimation of $\boldsymbol{\Theta}$ easy. However, we then lose the connection to the actual factorization; while the model is still parameterized through the low-rank matrices **U** and **V**, they become unidentifiable. In practice, the model can still be useful: If the goal is not to analyze the components but to find a low-rank approximation of **X** (which is sufficient e.g. for reconstructing the original data from a compressed version), then it is feasible to place the prior directly on $\boldsymbol{\Theta}$.

To combine the advantages of the above two formulations, we introduce the general prior family

$$p(\mathbf{U}, \mathbf{V}) = \frac{1}{Z(\boldsymbol{\psi})} a(\mathbf{UV})^\beta b(\mathbf{U})^\gamma c(\mathbf{V})^\gamma, \quad (3)$$

where $\beta$ and $\gamma$ are control parameters. The functions $a(\cdot)$, $b(\cdot)$, and $c(\cdot)$ can be arbitrary non-negative functions over the domain of the parameters, and $\boldsymbol{\psi}$ denotes collectively the parameters of all of them. The entire normalization is done with $Z(\boldsymbol{\psi})$ so the functions $a(\cdot)$, $b(\cdot)$ and $c(\cdot)$ need not be normalized. In practice, however, one would typically use simple standard distributions. Then $(\beta = 0, \gamma = 1)$ and $(\beta = 1, \gamma = 0)$ reduce the prior into the simpler alternatives discussed above, and setting $\gamma = 1 - \beta$ provides a single-parameter family for interpolating between the two.

A useful property of the prior is that if $a(\cdot)$ is set to give zero for values outside the domain of $\boldsymbol{\Theta}$, then already a small $\beta$ will be sufficient to restrict $b(\mathbf{U})^\gamma c(\mathbf{V})^\gamma$ to be a valid prior. More generally, $a(\mathbf{UV})$ can be thought of as a regularization term, making the model less sensitive for the specific choice of the distributions $b(\mathbf{U})$ and $c(\mathbf{V})$. In practice we simply use component-wise Gaussian priors,

$$b(\mathbf{U}) = \prod_{n=1}^{N} b(\mathbf{U}_{n,:}) = \prod_{n=1}^{N} N(\mathbf{0}, \boldsymbol{\Sigma}_{\mathbf{U}})$$
$$c(\mathbf{V}) = \prod_{k=1}^{K} c(\mathbf{V}_{k,:}) = \prod_{k=1}^{K} N(\mathbf{0}, \boldsymbol{\Sigma}_{\mathbf{V}_k}) \quad (4)$$

for both, which would not work in general without the regularizing $a(\mathbf{UV})$ term which we choose conjugate.

A practical challenge is that the prior is known only up to the normalization constant $Z$. While it cancels out when inferring the parameters, the unknown constant makes inference on hyper-parameters $\boldsymbol{\psi}$ of the prior difficult. We will discuss the solutions for this in the next section, separately for each inference algorithm.

## 5 INFERENCE

In principle the inference process for models of coupled data is identical to that of standard EPCA. The only difference between the models is the set of zeros in **V**, which requires only trivial modifications for most algorithms. In practice, however, there are certain challenges that need addressing. We will first recap standard inference methods for EPCA models in general, including details on how our new prior affects them, and then present a novel two-level sampling approach that solves some of the challenges. The details are given for the CCA variant; the other models are special cases of that. In the experiment section we provide examples for each of the inference algorithms.

### 5.1 POINT ESTIMATES

Point estimates of **U** and **V** can be inferred from data by maximizing the log likelihood that essentially measures the similarity between the data and the low-rank approximation. MAP estimation is conceptually equally simple; the priors only result in additive terms in the log-likelihood. Guo and Schuurmans (2008) proposed a convex optimization algorithm for the maximum-likelihood case, while MAP estimation requires more generic optimization algorithms. Following Srebro and Jaakkola (2003), we use conjugate gradients, which has in our experiments turned out to be sufficiently robust.

For MAP inference the hyperparameters of the prior $p(\mathbf{U}, \mathbf{V})$ are chosen by cross-validation. To avoid needing to validate over the Cartesian product of all of the parameters we choose $\gamma = 1 - \beta$ and use a simple approach where the hyperparameters of $a(\mathbf{UV})$ are chosen by assuming $\beta = 1$ and the hyperparameters of $b(\mathbf{U})$ and $c(\mathbf{V})$ by assuming $\beta = 0$. We show in the experiments section that already this simple approach leads to a better generalization ability than using either of the extremes, for a wide range of $\beta$.

### 5.2 HMC SAMPLER

For full Bayesian analysis Mohamed et al. (2009) applied a Hybrid Monte Carlo (HMC) sampler. Com-

pared to standard Metropolis-Hastings (MH) samplers, the HMC typically converges faster in large state spaces due to utilizing the gradient information.

Inferring the hyperparameters is difficult since the prior has an intractable normalization constant. We apply the exchange algorithm of Murray et al. (2006). The algorithm works with standard MH proposals for the hyperparameters $\psi$ but multiplies the acceptance probability by $f(\mathbf{U}^*, \mathbf{V}^*|\psi_t)/f(\mathbf{U}^*, \mathbf{V}^*|\psi_{t+1})$, where $\mathbf{U}^*$ and $\mathbf{V}^*$ are auxiliary variables or "replacement data" drawn from the prior using a separate MCMC chain for each posterior sample $\psi_t$, and $f(\mathbf{U}, \mathbf{V})$ equals (3) without the normalization term $Z(\psi)$. The algorithm was originally proposed for exact samples; we use randomly initialized MCMC chains sampled until convergence, resulting in an approximative variant.

### 5.3 ALTERNATING SAMPLER

Without further measures the ECCA model suffers from two kinds of unidentifiability problems, which makes inference difficult. First, the solution is defined only up to a rotation of $\mathbf{U}$ (as for EPCA in general). Second, it is hard to make sure the modeling power is divided correctly between the $\mathbf{U}_S$, $\mathbf{U}_1$, and $\mathbf{U}_2$. For Gaussian BCCA both problems can be solved, by analytically marginalizing the source-specific noise out and finding the right rotation (Klami and Kaski, 2007). Unfortunately, analytic marginalization is not possible for other exponential family distributions, which results in less efficient inference.

We next introduce a novel sampler that utilizes the more efficient solutions for Gaussian models as part of the sampler for general exponential families. The approach is similar to how Hoff (2007) made inference for binary PCA. The intuitive idea is to alternate between two sampling stages. In one stage, we treat the parameters $\Theta$ as data that *a priori* follows normal distribution, and learn a factorization $\Theta = \mathbf{UV}$ for that. The other stage then updates $\Theta$, taking into account both the exponential family likelihood and the conjugate prior $a(\mathbf{UV})$.

The practical sampling algorithm, coined GiBECCA, proceeds by alternating between two separate sampling steps implementing the above idea. Given the current sample for $\Theta$ we apply the Gibbs sampler for Gaussian BCCA (Klami and Kaski, 2007), treating $\Theta$ as data. This gives a new posterior sample for $\mathbf{U}_S$ and $\mathbf{V}$, as well as a block-diagonal noise covariance $\Sigma = [\mathbf{V}_1\mathbf{V}_1^T, \mathbf{0}; \mathbf{0}, \mathbf{V}_2\mathbf{V}_2^T]$ obtained by marginalizing out $\mathbf{U}_1$ and $\mathbf{U}_2$. During this step, we can also easily infer the hyperparameters of $b(\mathbf{U})$ and $c(\mathbf{V})$, since the full prior is the product of them and hence the normalization constant is tractable.

Next, we sample a new parameter matrix $\Theta^*$ given the data and the current values for the model parameters, using MH. The trick is to use the predictive distribution of the Gaussian BCCA as the proposal distribution for $\Theta$. It produces parameters that are approximately normal and for which the factorization can effectively be found, yet the likelihood part takes the true distribution correctly into account. In detail, the proposals are drawn independently for each data point $n$ from $N(\mathbf{U}_S^{(n,:)}\mathbf{V}, \Sigma)$. For each element $\Theta_{nd}^*$ the new value is then accepted with probability

$$\min\left(1, \frac{p(\mathbf{X}_{nd}|\Theta_{nd}^*)a(\Theta_{nd}^*)^\beta}{p(\mathbf{X}_{nd}|\Theta_{nd})a(\Theta_{nd})^\beta}\right),$$

which takes into account both the likelihood and the remaining part of the prior. Possible domain constraints are taken into account by always rejecting proposals leading to $a(\Theta^*) = 0$. The $a(\Theta)$ part can be interpreted directly as a regularizing term for which fixing the hyperparameters manually is the right solution. For example, for binary data we can choose $a(\Theta)$ as the symmetric beta distribution, which enables interpreting $\beta$ directly as the strength of the prior. Similarly, for count data we can use a gamma prior, where $\beta$ controls the variance of the counts.

## 6 EXPERIMENTS

### 6.1 SUPERVISED EPCA

The first empirical experiment shows the importance of separately modeling the data-specific noise in supervised learning. Using artificial toy data, we demonstrate the difference between supervised EPCA (SEPCA) and EPLS.

We created binary data from the model (2) with $K_S = 1$, $K_2 = 5$, $D_1 = 1$, and $D_2 = 20$. We used 50 samples for training and 950 for testing. We found the MAP estimate of the parameters with $\beta = 0$ and $\gamma = 1$ for the prior, and compared the models in the task of predicting $\mathbf{Y}_1$ for the left-out testing samples, using prediction error as the performance measure. The results were averaged over 80 random data sets.

As the shared source is only one-dimensional, it is possible to reach maximal prediction accuracy already with one component. However, SEPCA with just one component did not find the true solution as it is confused by the noise specific to $\mathbf{Y}_2$. The model will still reach the optimal prediction accuracy, but requires 6 components for it (Figure 2). The trick of Rish et al. (2008), lowering the importance of modeling $\mathbf{Y}_2$, helps by improving the predictive performance for low numbers of components, but still as many components are needed for optimal performance.

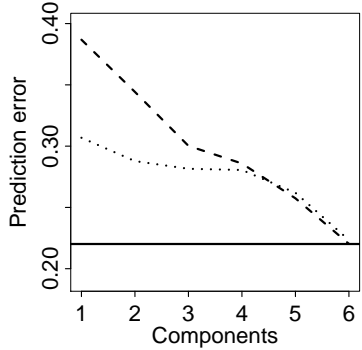

Figure 2: Prediction errors (lower is better) for the supervised EPCA experiment. The solid line depicts the error for a one-component EPLS-solution, while the other two curves are classical SEPCA models. The dashed line assumed equal modeling power for the target and covariates, while the dotted line weights the covariate modeling part with $\alpha = 10^{-3}$. A wide range of values result in similar performance (not shown).

EPLS, instead, found the true one-dimensional shared space, while modeling all the source-specific noise with separate components. Hence, it achieved the same predictive performance already with a single component, and for $1-5$ components it was significantly more accurate ($p < 0.05$, t-test with Bonferroni correction). It is worth noting, however, that the computational load for optimal prediction is comparable; SEPCA required 6 components, while EPLS required 1 shared and 5 noise components. The added benefit of EPLS is primarily in interpretation.

### 6.2 THE EFFECT OF THE PRIOR

In Section 4 we presented a family of prior distributions controlled by the regularization parameter $\beta$ (we set here $\gamma = 1 - \beta$). Here we illustrate how the combination improves the predictive performance of the model on the UCI SPECT data (http://archive.ics.uci.edu/ml/datasets/SPECT+Heart). We solve the standard PCA task of missing value imputation with one component for 100 values of the regularization parameter, and measure the performance as the reconstruction quality (log-likelihood).

As the data is binary, we choose the Bernoulli distribution and prior

$$a(\mathbf{UV}) = \prod_{n=1}^{N} \prod_{d=1}^{D} Beta(\lambda + 1, \nu - \lambda + 1).$$

with the computationally simple assumption of individual Gaussian priors for $\mathbf{U}$ and $\mathbf{V}$ as in (4) with $\mathbf{\Sigma_U} = \sigma_\mathbf{U}^2 \mathbf{I}$ and $\mathbf{\Sigma_{V_k}} = \sigma_\mathbf{V}^2 \mathbf{I}$ $\forall k$. We then find the MAP solution with conjugate gradients.

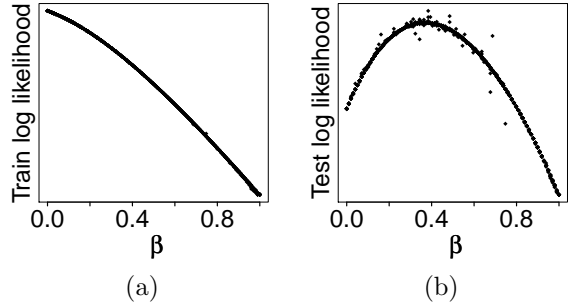

Figure 3: Illustration of the reconstruction quality with different values of the regularization parameter. For each $\beta$ we used 10 random initializations and included all of the results in the plot to illustrate that already the simple conjugate gradient algorithm always converges to the global optimum. The best generalization ability is obtained with $\beta \approx 0.4$, and the smooth curve indicates that the choice is robust and generalizes for further independent test sets (not shown).

The main purpose of the experiment is to illustrate the effect of the regularization parameter $\beta$. We first learn suitable values for the hyperparameters for $\beta = 0$ and $\beta = 1$ separately with simple cross-validation, resulting in values $\sigma_\mathbf{V}^2 = 100$, $\sigma_\mathbf{U}^2 = 0.001$, $\lambda = 0.1$, and $\nu = 0.2$. We then vary the $\beta$ parameter, keeping the hyperparameters fixed, and show (Figure 3) that the optimal predictive performance is obtained with $\beta$ around 0.4. That is, regularizing a model with separate priors for $\mathbf{U}$ and $\mathbf{V}$ by conjugate prior on $\mathbf{UV}$ improves the predictive performance.

### 6.3 EXPONENTIAL FAMILY CCA

#### 6.3.1 Classification in the joint space

CCA finds a shared representation that contains the variation to both data sources. The ability to do that can be indirectly measured by attempting to classify the samples given the shared representation. On an artificial data where the shared variation is known to be relevant (i.e., predictive of the class labels), a model extracting the true shared variation should have the best performance.

We created two collections of toy data sets from the model in Figure 1 with $K_S = 1$, $K_1 = 2$ and $K_2 = 2$. The first collection was binary and the second was count data. We chose $N = 50$ and $D_1 = D_2 = 20$, and learned four different variants of CCA for 10 randomly created data sets to study the effect of the link function and inference algorithm. First, we applied standard linear CCA to obtain a baseline. Bayesian Gaussian CCA (Klami and Kaski, 2007), which has an incorrect

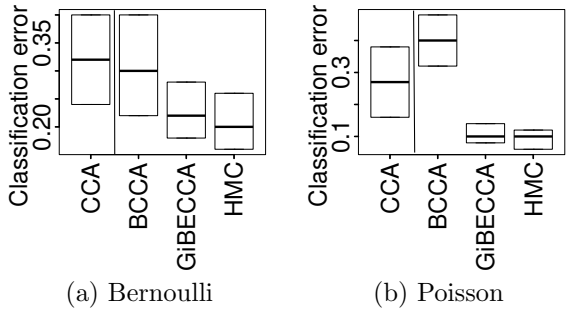

(a) Bernoulli  (b) Poisson

|  | BCCA | GiBECCA | HMC |
|---|---|---|---|
| Bernoulli | 0.54s | 0.70s | 6.85s |
| Poisson | 0.73s | 0.88s | 9.02s |

(c) Time between non-correlated samples

Figure 4: Performance of CCA variants, measured as the classification error of a K-nearest neighbor classifier ($K = 9$) in the shared latent space. For both Bernoulli (a) and Poisson (b) observations the two inference algorithms for exponential family CCA (GiBECCA and HMC) outperformed the Gaussian variant (BCCA) and standard CCA baseline (the boxplots show the 25%, 50% and 75% quantiles). The difference is particularly clear for the skewed Poisson distribution (b), where making the incorrect Gaussianity assumption even decreases the performance compared to classical CCA. GiBECCA and HMC have comparable accuracy for both data types, but the former is an order of magnitude faster, having only a small overhead over the Gaussian Gibbs sampler (c). The numbers show the mean CPU time between non-correlated posterior samples (autocorrelation below 0.1).

link function here (namely the identity function), is comparable with CCA on the binary data, but worse on the skewed count data (Figure 4).

The exponential family CCA with correct distributional assumptions outperformed the alternatives for both data sets. For binary data the standard models are reasonable but still worse than the exponential family variants, whereas for the count data the difference is considerable. We show results for both the HMC sampler and GiBECCA, using mild regularization with $\beta = 0.1$ for both. The accuracy of both inference methods is comparable for both data types, but GiBECCA is an order of magnitude more efficient, largely due to the inefficiency caused by inference of the hyperparameters in the HMC sampler.

#### 6.3.2 Movie data

To demonstrate the data analysis capabilities of Bayesian ECCA, we analyze a small collection of movies described with two views, selected from information available in the Allmovie database (http://www.allmovie.com/). The first view is the binary bag-of-words representation of a brief description of the movie, while the other is a multivariate genre classification in binary format. Each movie may belong to a subset of 10 genres, which extends the task beyond supervised visualization or SPCA.

Our main interest is in demonstrating the capability of ECCA to separate shared information from structured "noise" present in only one of the views. Hence, we manually construct the representation of the content descriptions to contain both. We manually choose a subset of terms (total of 32 terms) for the bag-of-words representation, so that half of the terms were chosen as genre-related and half were other terms chosen near the genre-related terms in frequency order to provide a contrast group. As an example, the most frequent terms in the genre-related set are *love*, *comedy* and *drama*, while the corresponding words in the noise set are *two*, *woman*, *some*, chosen because their frequency matched best the genre-related words.

We apply GiBECCA on this data, aiming to extract the components that best capture the genre variation. Figure 5 shows the first two shared projection vectors, that is, the first two rows of $\mathbf{V}$. We immediately see that the part covering the noise-terms in $\mathbf{V}_{S2}$ is close to zero for all terms, showing that the shared components do not capture description-specific noise. At the same time, each projection picks a subset of genre-related terms and actual genre memberships. Closer inspection of the features reveals that the first component separates romantic movies from action movies, while the second component mainly separates family-targeted genres (cartoons, family movies) from drama.

## 7 DISCUSSION

We presented a general framework for matrix factorizations or projection methods in the exponential family, and derived methods for analyzing coupled data sources. We also introduced a new family of prior distributions for the Bayesian analysis of EPCA models. We combine separate priors on the latent variables and projections, needed to make the solution identifiable and interpretable, with a regularizing prior specified directly for the natural parameters of the exponential family. As a result we can make computationally tractable assumptions for the latent variables while still getting a valid prior. The prior is known only up to a normalization constraint, but we show how it it still possible to infer even the hyperparameters of the prior. This is particularly efficient in our new sampler that uses a Gibbs sampler for the Gaussian

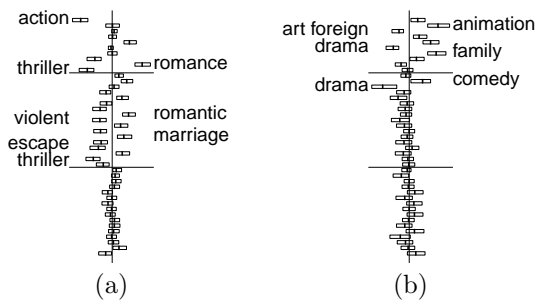

Figure 5: Illustration of the first two CCA components of the movie data. In both figures the top 10 bars represent the 10 genre membership indicators, the next 16 bars the genre-related words in the textual description of the movie, and the bottom 16 bars the genre-independent terms. Genre-related terms are present in the projections much more strongly than the genre-independent 'noise'-terms, as they should. The first shared component (a) picks most genre-related terms, detecting a strong link between the genre memberships and descriptions. The second shared component (b) extracts a more detailed relationship: family/comedy/animation movies are separated from the rest by absence of the word *drama* in the descriptions.

distribution to create proposals for the CCA model in any exponential family.

However, there is still work to be done, especially for the most flexible model corresponding to CCA. The novel efficient sampler explicitly marginalizes out the components specific to the individual data sources, but needs to use a normality assumption for that. With more flexible distributions for latent variables we need to represent also those components explicitly, which results in identifiability problems and requires a heavy exchange algorithm for hyperparameter inference. Computationally efficient algorithms for the most general case are hence still missing, but the experiments in this article suggest that the GiBECCA algorithm, making a partial assumption of normality, is a practical learning tool for exponential family CCA.

### Acknowledgements

The authors belong to Adaptive Information Research Centre, a CoE of Academy of Finland. The work was supported by Academy of Finland decision number 133818, and in part by the PASCAL2 EU NoE.